# Shallow Parsing Pipeline for Hindi-English Code-Mixed Social Media Text

**Arnav Sharma** and **Sakshi Gupta** and **Raveesh Motlani** and **Piyush Bansal**
and **Manish Shrivastava** and **Radhika Mamidi** and **Dipti M. Sharma**
Kohli Center on Intelligent Systems (KCIS)
International Institute of Information Technology, Hyderabad (IIIT Hyderabad)
Gachibowli, Hyderabad, Telangana 500032
{arnav.s, sakshi.gupta, raveesh.motlani, piyush.bansal}@research.iiit.ac.in
{m.shrivastava, radhika, dipti}@iiit.ac.in

## Abstract

In this study, the problem of shallow parsing of Hindi-English code-mixed social media text (CSMT) has been addressed. We have annotated the data, developed a language identifier, a normalizer, a part-of-speech tagger and a shallow parser. To the best of our knowledge, we are the first to attempt shallow parsing on CSMT. The pipeline developed has been made available to the research community with the goal of enabling better text analysis of Hindi English CSMT. The pipeline is accessible at [1].

## 1 Introduction

Multilingual speakers tend to exhibit code-mixing and code-switching in their use of language on social media platforms. Code-Mixing is the embedding of linguistic units such as phrases, words or morphemes of one language into an utterance of another language whereas code-switching refers to the co-occurrence of speech extracts belonging to two different grammatical systems (Gumperz., 1982). Here we use code-mixing to refer to both the scenarios.

Hindi-English bilingual speakers produce huge amounts of CSMT. Vyas et al. (2014) noted that the complexity in analyzing CSMT stems from non-adherence to a formal grammar, spelling variations, lack of annotated data, inherent conversational nature of the text and of course, code-mixing. Therefore, there is a need to create datasets and Natural Language Processing (NLP) tools for CSMT as traditional tools are ill-equipped for it. Taking a step in this direction, we describe the shallow parsing pipeline built during this study.

## 2 Background

Bali et al. (2014) gathered data from Facebook generated by English-Hindi bilingual users which on analysis, showed a significant amount of code-mixing. Barman et al. (2014) investigated language identification at word level on Bengali-Hindi-English CSMT. They annotated a corpus with more than 180,000 tokens and achieved an accuracy of 95.76% using statistical models with monolingual dictionaries.

Solorio and Liu (2008) experimented with POS tagging for English-Spanish Code-Switched discourse by using pre-existing taggers for both languages and achieved an accuracy of 93.48%. However, the data used was manually transcribed and thus lacked the problems added by CSMT. Vyas et al. (2014) formalized the problem, reported challenges in processing Hindi-English CSMT and performed initial experiments on POS tagging. Their POS tagger accuracy fell by 14% to 65% without using gold language labels and normalization. Thus, language identification and normalization are critical for POS tagging (Vyas et al., 2014), which in turn is critical further down the pipeline for shallow parsing as evident in Table 5.

Jamatia et al. (2015) also built a POS tagger for Hindi-English CSMT using Random Forests on 2,583 utterances with gold language labels and achieved an accuracy of 79.8%. In the monolin-gual social media text context, Gimpel et al. (2011) built a POS tagger for English tweets and achieved an accuracy of 89.95% on 1,827 annotated tweets. Owoputi et al. (2013) further improved this POS tagger, increasing the accuracy to 93%.

---

[1] http://bit.ly/csmt-parser-api

| Lang. | Sentences |
|---:|:---:|
| English | 141 (16.43%) |
| Hindi | 111 (12.94%) |
| Code-mixed | 606 (70.63%) |
| Total | 858 |

Table 1: Data distribution at sentence level.

| Lang. | All Sentences | Only CM Sentences |
|---|---|---|
| Hindi | 6318 (57.05%) | 5352 (63.34%) |
| English | 3015 (27.22%) | 1886 (22.32%) |
| Rest | 1742 (15.73%) | 1212 (14.34%) |
| Total | 11075 | 8450 |

Table 2: Data distribution at token level.

## 3 Data Preparation

CSMT was obtained from social media posts from the data shared for Subtask 1 of FIRE-2014 Shared Task on Transliterated Search. The existing annotation on the FIRE dataset was removed, posts were broken down into sentences and 858 of those sentences were randomly selected for manual annotation.

Table 1 and Table 2 show the distribution of the dataset at sentence and token level respectively. The language of 63.33% of the tokens in code-mixed sentences is Hindi. Based on the distribution, it is reasonable to assume that Hindi is the matrix language (Azuma, 1993; Myers-Scotton, 1997) in most of the code-mixed sentences.

### 3.1 Dataset examples

1. hy... try fr sm gov job jiske forms niklte h...
   **Gloss:** Hey... try for some government job which forms give out...
   **Translation:** Hey... try for some government job which gives out forms...

2. To tum divya bharti mandir marriage kendra ko donate karna
   **Gloss:** So you divya bharti temple marriage center to donate do
   **Translation:** So you donate to divya bharti temple marriage center

The dataset is comprised of sentences similar to example 1 and 2. Example 1 shows code-switching as the language switches from English to Hindi whereas example 2 shows code-mixing as some English words are embedded in a Hindi utterance. Spelling variations (`sm` - `some`, `gov` - `government`), ambiguous words (`To` - `So` in Hindi or `To` in English) and non-adherence to a formal grammar (out of place ellipsis - `...`, no or misplaced punctuation) are some of the challenges evident in analyzing the examples above.

### 3.2 Annotation

Annotation was done on the following four layers:

1. **Language Identification**: Every word was given a tag out of three 'en', 'hi' and 'rest' to mark its language. Words that a bilingual speaker could identify as belonging to either Hindi or English were marked as 'hi' or 'en'. The label 'rest' was given to symbols, emoticons, punctuation, named entities, acronyms, foreign words and words with sub-lexical code-mixing like `chapattis` (Gloss: chapatti - bread) which is a Hindi word (*chapatti*) following English morphology (plural marker -*s*).

2. **Normalization**: Words with language tag 'hi' in Roman script were labeled with their standard form in the native script of Hindi, Devanagari. Similarly, words with language tag 'en' were labeled with their standard spelling. Words with language tag 'rest' were kept as they are. This acted as testing data for our Normalization module.

3. **Parts-of-Speech (POS)**: Universal POS tagset (Petrov et al., 2011) was used to label the POS of each word as this tagset is applicable to both English and Hindi words. Sub-lexical code-mixed words were annotated based on their context, since POS is a function of a word in a given context. For example, an English verb used as a noun in Hindi context is labeled as a noun.

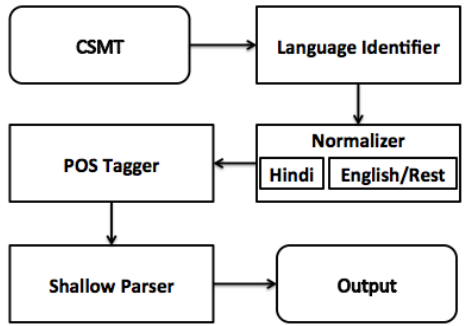

Figure 1: Schematic Diagram of the Pipeline

| Features | Accuracy |
|---|---|
| BNC | 61.26 |
| +LEXNORM | 71.43 |
| +HINDI_DICT | 77.50 |
| +NGRAM | 93.18 |
| +AFFIXES | 93.98 |

Table 3: Feature Ablation for Language Identifier

4. **Chunking**: A chunk tag comprises of chunk label and chunk boundary. The chunk label tagset is a coarser version of AnnCorra tagset (Bharati et al., 2006). Unlike AnnCorra, only one tag is used for all verb chunks in our tagset. Chunk boundary is marked using BI notation where 'B-' prefix indicates beginning of a chunk and 'I-' prefix indicates that the word is inside a chunk.

This whole dataset was annotated by eight Hindi-English bilingual speakers. Two other annotators reviewed and cleaned it. To measure inter-annotator agreement, another annotator read the guidelines and annotated 25 sentences (334 tokens) from scratch. The inter-annotator agreement calculated using Cohen's $\kappa$ (Cohen, 1960) came out to be 0.97, 0.83 and 0.89 for language identification, POS tagging and shallow parsing respectively.

## 4 Shallow Parsing Pipeline

Shallow parsing is the task of identifying and segmenting text into syntactically correlated word groups (Abney, 1992; Harris, 1957). Shallow parsing is a viable alternative to full parsing as shown by (Li and Roth, 2001). Our shallow parsing pipeline is composed of four main modules, as shown in Figure 1. These modules, in the order of their usage, are *Language Identification*, *Normalization*, *POS Tagger* and *Shallow Parser*.

Our pipeline takes a raw utterance in Roman script as input on which each module runs sequentially. Twokenizer[2] (Owoputi et al., 2013) which performs well on Hindi-English CSMT (Jamatia et al., 2015) was used to tokenize the utterance into words. The *Language Identification* module assigns each token a language label. Based on the language label assigned, the *Normalizer* runs the Hindi normalizer or the English/Rest normalizer. The *POS tagger* uses the output of the normalizer to assign each word a POS tag. Finally, the *Shallow Parser* assigns a chunk label with boundary.

[2] http://www.ark.cs.cmu.edu/TweetNLP/

The functionality and performance of each module is described in greater detail in the following subsections.

### 4.1 Language Identification

While language identification at the document level is a well-established task (McNamee, 2005), identifying language in social media posts has certain challenges associated to it. Spelling errors, phonetic typing, use of transliterated alphabets and abbreviations combined with code-mixing make this problem interesting. Similar to (Barman et al., 2014), we performed two experiments treating language identification as a three class ('hi', 'en', 'rest') classification problem. The feature set comprised of - **BNC**: normalized frequency of the word in British National Corpus (BNC)[3]. **LEXNORM**: binary feature indicating presence of the word in the lexical normalization dataset released by Han et al. (2011). **HINDI_DICT**: binary feature indicating presence of the word in a dictionary of 30,823 transliterated Hindi words as released by Gupta (2012). **NGRAM**: word n-grams. **AFFIXES**: prefixes and suffixes of the word.

Using these features and introducing a context-window of n-words, we trained a linear SVM. In another experiment we modeled language identification as a sequence labeling task, where we employed CRF into usage. The idea behind this was that code-mixed text has some inherent structure which is largely dictated by the matrix language of the text. The latter approach using CRF had a greater accuracy, which validated our hypothesis. The results of this module are shown in Table 3.

[3] http://www.natcorp.ox.ac.uk/

### 4.2 Normalization

Once the language identification task was complete, there was a need to convert the noisy non-standard tokens (such as Hindi words inconsistently written in many ways using the Roman script) in the text into standard words. To fix this, a normalization module that performs language-specific transformations, yielding the correct spelling for a given word was built. Two language specific normalizers, one for Hindi and other for English/Rest, had two subnormalizers each, as described below. Both subnormalizers generated normalized candidates which were then ranked, as explained later in this subsection.

1. **Noisy Channel Framework**: A generative model was trained to produce noisy (unnormalized) tokens from a given normalized word. Using the model's confidence score and the probability of the normalized word in the background corpus, *n*-best normalizations were chosen. First, we obtained character alignments between noisy Hindi words in Roman script ($H_r$) to normalized Hindi words-format($H_w$) using GIZA++ (Och and Ney, 2003) on 30,823 Hindi word pairs of the form ($H_w$ - $H_r$) (Gupta et al., 2012). Next, a CRF classifier was trained over these alignments, enabling it to convert a character sequence from Roman to Devanagari using learnt letter transformations. Using this model, noisy $H_r$ words were created for $H_w$ words obtained from a dictionary of 1,17,789 Hindi words (Biemann et al., 2007). Finally, using the formula below, we computed the most probable $H_w$ for a given $H_r$.

$$H_w = argmax_{H_{w_i}} p(H_{w_i}|H_r)$$
$$= argmax_{H_{w_i}} p(H_r|H_{w_i}) p(H_{w_i})$$

   where $p(H_{w_i})$ is the probability of word $H_{w_i}$ in the background corpus.

| Features | Accuracy |
|---|---|
| Baseline | 69.27 |
| +LANG | 70.44 |
| +NORM | 72.61 |
| +TPOS | 73.18 |
| +HPOS, -TPOS | 73.55 |
| +COMBINED | 75.07 |

**Table 4:** Feature Ablation for POS Tagger

2. **SILPA Spell Checker**: This subnormalizer uses SILPA libindic spell-checker[4] to compute the top 10 normalized words for a given input word.

The candidates obtained from these two systems are ranked on the basis of the observed precision of the systems. The top-k candidates from each system are selected if they have a confidence score greater than an empirically observed Λ. A similar approach was used for English text normalization, using the English normalization pairs from (Han et al., 2012) and (Liu et al., 2012) for the noisy channel framework, and Aspell[5] as the spell-checker. Words with language tag 'rest' were left unprocessed. The accuracy for the Hindi Normalizer was 78.25%, and for the English Normalizer was 69.98%. The overall accuracy of this module is 74.48%; P@n (Precision@n) for n=3 is 77.51% and for n=5 is 81.76%.

### 4.3 Part-Of-Speech Tagging

Part-of-Speech (POS) tagging provides basic level of syntactic analysis for a given word or sentence. It was modeled as a sequence labeling task using CRF. The feature set comprised of - **Baseline**: Word based features - affixes, context and the word itself. **LANG**: Language label of the token. **NORM**: Normalized lexical features. **TPOS**: Output of Twitter POS tagger (Owoputi et al., 2013). **HPOS**: Output of IIIT's Hindi POS tagger[6]. **COMBINED**: HPOS for Hindi words and TPOS for English and Rest. The results of POS Tagger are shown in Table 4.

---

[4] https://github.com/libindic/spellchecker
[5] http://aspell.net/
[6] http://ltrc.iiit.ac.in/showfile.php?filename=downloads/shallow_parser.php

| Features | L | B | C |
|---|---|---|---|
| POS Tag | 88.01 | 78.75 | 76.64 |
| +POS Context [W5] | 87.92 | 81.36 | 78.09 |
| +POS_LEX | 88.18 | 81.46 | 78.58 |
| +NORMLEX | 88.25 | 82.17 | 78.73 |

Table 5: Feature Ablation for Shallow Parser

| | | P1 | P2 | E |
|---|---|---|---|---|
| LI | | 93.98 | 93.98 | NA |
| Norm | | 70.32 | 74.48 | 4.16 |
| POS | | 68.25 | 75.07 | 6.82 |
| | L | 75.73 | 88.25 | 12.52 |
| SP | B | 74.96 | 82.17 | 7.21 |
| | C | 61.95 | 78.73 | 16.78 |

Table 6: Pipeline accuracy and error propagation. LI = Language Identification, Norm = Normalizer, POS = POS Tagger, SP = Shallow Parser, L = Label, B = Boundary, C = Combined, P1 = Actual Pipeline, P2 = Gold Pipeline, E = Error Propagation

### 4.4 Shallow Parsing

A chunk comprises of two aspects - the chunk boundary and the chunk label. Shallow Parsing was modeled as three separate sequence labeling problems: **Label**, **Boundary** and **Combined**, for each of which a CRF model was trained. The feature set comprised of - **POS**: POS tag of the word. **POS Context**: POS tags in the context window of length 5, i.e., the two previous tags, current tag and next two tags. **POS_LEX**: A special feature made up of concatenation of POS and LEX. **NORMLEX**: The word in its normalized form. The results of this module are shown in Table 5.

## 5 Pipeline Results

The best performing model was selected from each module and was used in the pipeline. Table 6 tabulates the step by step accuracy of the pipeline calculated using 10 fold cross-validation.

## 6 Conclusion and Future Work

In this study, we have developed a system for Hindi-English CSMT data that can identify the language of the words, normalize them to their standard forms, assign them their POS tag and segment them into chunks. We have released the system.

In the future, we intend to continue creating more annotated code-mixed social media data. We would also like to improve upon the challenging problem of normalization of monolingual social Hindi sentences. Also, we would further extend our pipeline and build a full parser which has aplenty applications in NLP.

## References

Steven P Abney. 1992. *Parsing by chunks*. Springer.

Shoji Azuma. 1993. The frame-content hypothesis in speech production: Evidence from intrasentential code switching. *Linguistics*, 31(6):1071–1094.

Kalika Bali, Jatin Sharma, Monojit Choudhury, and Yogarshi Vyas. 2014. i am borrowing ya mixing ? an analysis of english-hindi code mixing in facebook. In *Proceedings of the First Workshop on Computational Approaches to Code Switching*, pages 116–126, Doha, Qatar, October. Association for Computational Linguistics.

Utsab Barman, Amitava Das, Joachim Wagner, and Jennifer Foster. 2014. Code mixing: A challenge for language identification in the language of social media. *EMNLP 2014*, page 13.

Akshar Bharati, Rajeev Sangal, Dipti Misra Sharma, and Lakshmi Bai. 2006. Anncorra: Annotating corpora guidelines for pos and chunk annotation for indian languages. *LTRC-TR31*.

Chris Biemann, Gerhard Heyer, Uwe Quasthoff, and Matthias Richter. 2007. The leipzig corpora collection-monolingual corpora of standard size. *Proceedings of Corpus Linguistic*.

Jacob Cohen. 1960. A coefficient of agreement for nominal scales. *Educational and Psychological Measurement*, 134:3746.

Kevin Gimpel, Nathan Schneider, Brendan O'Connor, Dipanjan Das, Daniel Mills, Jacob Eisenstein, Michael Heilman, Dani Yogatama, Jeffrey Flanigan, and Noah A Smith. 2011. Part-of-speech tagging for twitter: Annotation, features, and experiments. In *Proceedings of the 49th Annual Meeting of the Association for Computational Linguistics: Human Language Technologies: short papers-Volume 2*, pages 42–47. Association for Computational Linguistics.

John J. Gumperz. 1982. *Discourse Strategies*. Oxford University Press.

Kanika Gupta, Monojit Choudhury, and Kalika Bali. 2012. Mining hindi-english transliteration pairs from online hindi lyrics. In *LREC*, pages 2459–2465.

Bo Han and Timothy Baldwin. 2011. Lexical normalisation of short text messages: Makn sens a# twitter. In *Proceedings of the 49th Annual Meeting of the Association for Computational Linguistics: Human Lan-

*guage Technologies-Volume 1*, pages 368–378. Association for Computational Linguistics.

Bo Han, Paul Cook, and Timothy Baldwin. 2012. Automatically constructing a normalisation dictionary for microblogs. In *Proceedings of the 2012 joint conference on empirical methods in natural language processing and computational natural language learning*, pages 421–432. Association for Computational Linguistics.

Zellig S Harris. 1957. Co-occurrence and transformation in linguistic structure. *Language*, pages 283–340.

Anupam Jamatia, Björn Gambäck, and Amitava Das. 2015. Part-of-speech tagging for code-mixed english-hindi twitter and facebook chat messages. *Proceedings of Recent Advances in Natural Language Processing*, page 239.

Xin Li and Dan Roth. 2001. Exploring evidence for shallow parsing. In *Proceedings of the 2001 workshop on Computational Natural Language Learning-Volume 7*, page 6. Association for Computational Linguistics.

Fei Liu, Fuliang Weng, and Xiao Jiang. 2012. A broad-coverage normalization system for social media language. In *Proceedings of the 50th Annual Meeting of the Association for Computational Linguistics: Long Papers-Volume 1*, pages 1035–1044. Association for Computational Linguistics.

Paul McNamee. 2005. Language identification: a solved problem suitable for undergraduate instruction. *Journal of Computing Sciences in Colleges*, 20(3):94–101.

Carol Myers-Scotton. 1997. *Duelling languages: Grammatical structure in codeswitching*. Oxford University Press.

Franz Josef Och and Hermann Ney. 2003. A systematic comparison of various statistical alignment models. *Computational Linguistics*, 29(1):19–51.

Olutobi Owoputi, Brendan O'Connor, Chris Dyer, Kevin Gimpel, Nathan Schneider, and Noah A Smith. 2013. Improved part-of-speech tagging for online conversational text with word clusters. Association for Computational Linguistics.

Slav Petrov, Dipanjan Das, and Ryan McDonald. 2011. A universal part-of-speech tagset. *arXiv preprint arXiv:1104.2086*.

Thamar Solorio and Yang Liu. 2008. Part-of-speech tagging for english-spanish code-switched text. In *Proceedings of the Conference on Empirical Methods in Natural Language Processing*, pages 1051–1060. Association for Computational Linguistics.

Yogarshi Vyas, Spandana Gella, Jatin Sharma, Kalika Bali, and Monojit Choudhury. 2014. Pos tagging of english-hindi code-mixed social media content. In *Proceedings of the First Workshop on Codeswitching, EMNLP*.